\documentclass[11pt]{article}

\usepackage[final]{acl}

\usepackage{times}
\usepackage{latexsym}

\usepackage[T1]{fontenc}

\usepackage[utf8]{inputenc}
\usepackage{comment}
\usepackage{microtype}

\usepackage{inconsolata}
\usepackage{amsmath}
\usepackage{longtable}
\usepackage{graphicx}
\usepackage{tikz}
\usetikzlibrary{arrows.meta, positioning, shapes.geometric}
\usepackage{booktabs}
\usepackage{multirow}
\usepackage{fvextra}

\DefineVerbatimEnvironment{prompt}{Verbatim}{
    breaklines=true,
    breakanywhere=true,
    fontsize=\small
}
\title{Translating the Translator: Decomposing the Cost of English-Forced Inter-Agent Communication}

\author{
Kushagra Agrawal \\
Faculty of Science and\\
Engineering \\
Åbo Akademi University \\
Turku, Finland \\
\texttt{kushagra.agrawal@ieee.org}
\And
Yuming Feng \\
School of Computer Science and\\
Engineering \\
Chongqing Sanxia University of\\
Science and Technology \\
Chongqing, China \\
\texttt{ymfeng@sanxiau.edu.cn}
\And
Man-Fai Leung \\
School of Computing and\\
Information Science \\
Faculty of Science and\\
Engineering \\
Anglia Ruskin University \\
Cambridge, UK \\
\texttt{man-fai.leung@aru.ac.uk}\\
}

\begin{document}
\maketitle
\begin{abstract}
Multi-agent LLM architectures, such as LangChain and AutoGen, largely assume English as the lingua franca for internal inter-agent communication, even when the end-user task is non-English. We fill this gap by evaluating a two-agent extraction-answer core, with an additional back-translation agent in the English-forced condition, across four typologically diverse languages (Hindi, Chinese, Spanish, Arabic; $n=300$ per language) using the Aya-23-8B model. We compare a native-language pipeline to an English-forced one (which incorporates a final back-translation step from English to the user's language). We discover a statistically significant English-Forcing Tax (surviving a strict Bonferroni correction) that isolates the cost of English routing from general multi-agent orchestration overhead. Forcing inter-agent communication through English reduces Exact Match accuracy by 13.0 percentage points (Spanish) up to 30.6 percentage points (Hindi) compared to native-language multi-agent execution. Using chrF scores as a diagnostic measure of English-reference lexical overlap, we find that lower overlap is strongly associated with pipeline failure, consistent with translation loss being an important contributor to the observed performance drop. These findings suggest a compelling case for native-language routing in agent frameworks when the source and target languages are typologically distant, reducing a compounding translation tax.
\end{abstract}

\section{Introduction}

The LLM internal ``language of thought'' used by agent frameworks is primarily English, and current paradigms such as LangChain, AutoGen, and CrewAI are heavily biased towards this. More often than not, prompts, tool schemas, and intermediate natural-language handoffs between agents are hardcoded in English when a user submits a query in Hindi, Arabic, or Spanish. The underlying engineering rationale is that LLMs are mostly trained on English data, so their reasoning and logic capabilities are strongest in English, and therefore translating non-English inputs into English under the hood should lead to better results. The theoretical justification for this architectural decision comes from recent mechanistic interpretability work showing that multilingual transformers pass through an English-dominant latent region when reasoning in other languages \cite{wendler-etal-2024-llamas, zhao2024how}.

However, this engineering default neglects a widely recognized phenomenon: explicit cross-lingual translation is lossy. This risk is identified in two bodies of literature. To begin with, work on cross-lingual transfer shows that routing through an English-centric concept space induces a ``cross-lingual tax'' in LLMs, with performance degrading as linguistic distance from English increases \cite{artetxe-etal-2020-cross, bafna2024evaluating}. Research on multi-agent systems further shows that natural language serves as a lossy bottleneck for information transfer between LLMs, where token sampling discards information and hallucinations accumulate as the length of agent chains increases \cite{pham2024cipher, tang2025augmenting}. Even more broadly, recent work has shown that architectural decisions in agentic pipelines incur measurable performance costs not accounted for by the language model itself. The overhead of coordination and cumulative information loss at test time limit the returns from increased complexity in orchestration \citep{agrawal2026token}.

Although these risks are known, no previous research has isolated the intersection of these two fields. Whether forcing an agent pipeline through a single English translation bottleneck (forward translation, English reasoning, and back-translation) compounds cross-lingual loss compared to native routing remains untested. If such a loss exists, the stage of the pipeline that primarily drives it also remains an open question.

To this end, we design a tightly coupled two-agent extraction-answer core, consisting of an Extractor and an Answerer, with an additional back-translation agent in the English-forced condition, operating on parallel multilingual QA data (XQuAD). We evaluate three settings: a Single-Agent baseline (S), a Native Multi-Agent pipeline (N), and an English-Forced Multi-Agent pipeline (E), which routes inter-agent handoffs through English and back-translates the final output to the native language. We evaluate the pipeline on consumer hardware using a pre-trained multilingual model (Aya-23-8B) across four typologically diverse source languages: Hindi, Chinese, Spanish, and Arabic.

Our findings suggest that the cost of the English-internals default is both substantial and measurable. Compared to an otherwise identical native-language multi-agent pipeline, forcing inter-agent communication through English incurs a statistically significant English-Forcing Tax of up to 30.6 percentage points in Exact Match accuracy, with the magnitude of the penalty increasing with typological distance from English. We further perform a diagnostic correlational analysis using chrF as a diagnostic measure of English-reference lexical overlap, showing that this degradation is strongly associated with the forward-translation step reducing lexical overlap.

Our contributions are threefold:
\begin{itemize}

\item \textbf{Controlled Isolation Design:} We propose a 3-condition experimental design to isolate the architectural penalty of English-forced handoffs from the baseline cross-lingual weakness of the underlying model.

\item \textbf{Quantitative Isolation of the English-Forcing Tax:} By directly comparing native-language handoffs ($N$) with English-forced handoffs ($E$), we isolate an incremental accuracy drop of up to 30.6 percentage points that increases with typological distance from English, independent of baseline multi-agent orchestration overhead.

\item \textbf{Diagnostic Analysis:} We analyze chrF as a diagnostic measure of English-reference lexical overlap to diagnose pipeline failures, identifying forward-translation degradation as a strong correlate of downstream error that can be targeted through engineering interventions.

\end{itemize}

\section{Related Work}

Our study sits at the intersection of cross-lingual LLM degradation, latent English representations, and lossy multi-agent communication. 

\textbf{Cross-Lingual Degradation and Latent English.}
Previous research has extensively reported that LLMs underperform in non-English reasoning. Models demonstrate emerging multilingual chain-of-thought capabilities when prompted in the native language \cite{shi2023language}, but still tend to exhibit internal representations that favor English. Multilingual transformers have been shown to pass through an English-dominant latent region before reaching the target language \cite{wendler-etal-2024-llamas}. This mapping is modeled as a three-stage process through a shared concept space \cite{zhao2024how}, although models that are not anchored to English have been shown to activate distinct latent languages \cite{zhong-etal-2025-language}. This literature explains why practitioners default to English for agent internals, as it mirrors the model's inherent latent behavior. Cross-lingual benchmarks such as XQuAD \cite{artetxe-etal-2020-cross} further reveal that these performance gaps remain prevalent, and analyses of language variation demonstrate that this degradation scales with linguistic distance \cite{bafna2024evaluating}.

\textbf{Lossy Multi-Agent Communication.}
In parallel, a growing body of research has examined information loss in multi-agent LLM pipelines. Communication between software systems has traditionally been achieved through protocols such as HTTP and gRPC, although frameworks such as AutoGen \cite{wu2024autogen} implement inter-agent communication through natural-language message passing, and sequential pipelines such as Chain-of-Agents pass context along a chain of workers in the form of text \cite{zhang2024chain}. However, \cite{pham2024cipher} showed that natural-language communication between agents imposes a token sampling step in which information about the model's full belief distribution is lost, motivating a transition to raw embedding communication (CIPHER). Subsequent work has proposed richer protocols, such as state delta encoding, to capture a larger portion of an agent's reasoning trace \cite{tang2025augmenting}, and has studied how erroneous information propagates differently depending on the communication topology \cite{shen2025understanding}. This literature establishes natural-language handoffs as a known lossy baseline in multi-agent systems, independent of language.

\textbf{Equity and Motivation.}
The impact of this lossiness is not evenly distributed. Recent work by \cite{joshi-etal-2020-state} quantified the disparity in NLP resource allocation, showing that language technology remains heavily concentrated in a small number of high-resource languages. An orchestration penalty that scales with distance from English is therefore a concrete example of a well-established structural bias in the design of language technology, disproportionately disadvantaging typologically distant and lower-resource languages.

\textbf{The Gap.}
Previous work has separately shown that (i) NLP resources are heavily biased toward high-resource languages \cite{joshi-etal-2020-state}, (ii) LLMs implicitly use English as a pivot representation even when reasoning in other languages \cite{wendler-etal-2024-llamas, zhao2024how, zhong-etal-2025-language}, and (iii) natural-language communication is inherently lossy regardless of language, a problem actively studied in frameworks such as Chain-of-Agents \cite{zhang2024chain} and addressed through alternative communication protocols such as CIPHER \cite{pham2024cipher} and state-delta encoding \cite{tang2025augmenting}. We isolate the effect of forcing a multi-agent system, built following the natural-language conversational structure typical of frameworks such as AutoGen \cite{wu2024autogen}, to combine (ii) and (iii): explicit, repeated cross-lingual translation instead of an implicit, single-pass internal pivot.

\section{Methodology}

To disassociate the architectural overhead of English-based handoffs, we build a strictly controlled two-agent extraction-answer core, with an additional back-translation agent in the English-forced condition, running on parallel multilingual QA data. 

\subsection{Data and Model}

We evaluate on the XQuAD (Cross-lingual Question Answering Dataset) benchmark \cite{artetxe-etal-2020-cross}, which consists of professional human translations of SQuAD v1.1 questions and their respective contexts across 11 languages. Due to the requirements of robust statistical testing, we sample exactly 300 questions per language from a subset of four typologically diverse languages (es: Spanish; ar: Arabic; hi: Hindi; zh: Chinese). This sample size is large enough to provide a sufficient number of discordant pairs for robust statistical testing while remaining computationally feasible on consumer hardware.

For the underlying LLM, we use \texttt{aya-23-8b-8bit} \cite{aryabumi2024aya}, a state-of-the-art (SOTA) open-weight model explicitly fine-tuned for multilingual tasks. We intentionally select a powerful multilingual model so that the observed failures are indicative of an architectural bottleneck in the pipeline rather than arising from a weak base model. Hence, the reported penalties are likely an underestimate; models with weaker multilingual capabilities, particularly those that are more English-centric, would likely exhibit even steeper performance degradation.

\subsection{The Three Conditions}

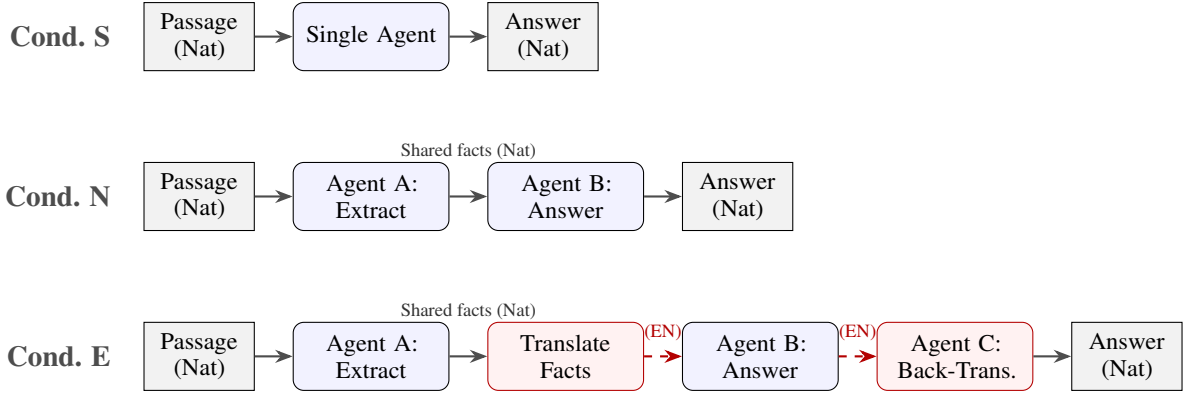
\begin{figure*}[t]
\centering
\begin{tikzpicture}[
    node distance=0.6cm and 0.5cm,
    agent/.style={rectangle, draw=black, rounded corners, minimum height=0.9cm, text width=1.8cm, align=center, fill=blue!5, font=\small},
    trans_agent/.style={rectangle, draw=red!70!black, rounded corners, minimum height=0.9cm, text width=1.8cm, align=center, fill=red!5, font=\small},
    data/.style={rectangle, draw=black, minimum height=0.9cm, text width=1.2cm, align=center, fill=gray!10, font=\small},
    normal/.style={->, >=Stealth, thick, color=black!70},
    trans/.style={->, >=Stealth, thick, dashed, color=red!70!black},
    sharedlabel/.style={midway, above=0.35cm, font=\scriptsize, text=black!80},
    enlabel/.style={midway, above=2pt, font=\scriptsize, text=red!70!black}
]

\node[data] (pass_s) {Passage\\(Nat)};
\node[agent, right=of pass_s] (agent_s) {Single Agent};
\node[data, right=of agent_s] (ans_s) {Answer\\(Nat)};

\draw[normal] (pass_s) -- (agent_s);
\draw[normal] (agent_s) -- (ans_s);
\node[left=0.3cm of pass_s, text=black!70, anchor=east, font=\bfseries] {Cond. S};

\node[data, below=1.2cm of pass_s] (pass_n) {Passage\\(Nat)};
\node[agent, right=of pass_n] (agent_a) {Agent A:\\Extract};
\node[agent, right=of agent_a] (agent_b_n) {Agent B:\\Answer};
\node[data, right=of agent_b_n] (ans_n) {Answer\\(Nat)};

\draw[normal] (pass_n) -- (agent_a);
\draw[normal] (agent_a) -- (agent_b_n) node[sharedlabel] {Shared facts (Nat)};
\draw[normal] (agent_b_n) -- (ans_n);
\node[left=0.3cm of pass_n, text=black!70, anchor=east, font=\bfseries] {Cond. N};

\node[data, below=1.2cm of pass_n] (pass_e) {Passage\\(Nat)};
\node[agent, right=of pass_e] (agent_a_e) {Agent A:\\Extract};
\node[trans_agent, right=of agent_a_e] (trans_a) {Translate\\Facts};
\node[agent, right=of trans_a] (agent_b_e) {Agent B:\\Answer};
\node[trans_agent, right=of agent_b_e] (agent_c_e) {Agent C:\\Back-Trans.};
\node[data, right=of agent_c_e] (ans_e) {Answer\\(Nat)};

\draw[normal] (pass_e) -- (agent_a_e);
\draw[normal] (agent_a_e) -- (trans_a) node[sharedlabel] {Shared facts (Nat)};
\draw[trans] (trans_a) -- (agent_b_e) node[enlabel] {(EN)};
\draw[trans] (agent_b_e) -- (agent_c_e) node[enlabel] {(EN)};
\draw[normal] (agent_c_e) -- (ans_e);

\node[left=0.3cm of pass_e, text=black!70, anchor=east, font=\bfseries] {Cond. E};

\end{tikzpicture}

\caption{Pipeline architecture for the three experimental conditions. In the multi-agent setting, Agent~A performs a single extraction of relevant facts in the native language. The same extracted facts are used by both Conditions N and E. Condition N directly answers using the native-language facts, whereas Condition E first translates the shared facts into English, answers in English using the aligned English question, and finally back-translates the answer into the native language. Thus, the only architectural difference between Conditions N and E is the introduction of the English-pivot pathway, comprising forward translation, English-language processing, and back-translation (dashed red arrows).}

\label{fig:pipeline}
\end{figure*}

We specifically evaluate a multi-agent setup rather than a single-agent translation baseline because our objective is to critique the default inter-agent communication protocols of frameworks like AutoGen and LangChain, which inherently rely on multi-agent message passing. For each question, the pipeline is executed under three different settings (illustrated in Figure~\ref{fig:pipeline}):

\begin{itemize}

\item \textbf{Condition S (Single-Agent Baseline):} A single agent reads the passage in the native language and directly answers the question in the native language. This establishes the baseline cross-lingual performance of the model without any multi-agent handoff.

\item \textbf{Condition N (Native Multi-Agent):} Agent A (Extractor) reads the passage in the native language and extracts relevant facts. Agent B (Answerer) receives \textit{only} the extracted facts from Agent A and produces the final answer in the native language.

\item \textbf{Condition E (English-Forced Multi-Agent):} Agent A reads the passage in the native language, extracts the relevant facts, and translates them into English. Agent B receives only the English facts and produces an answer in English. Finally, as an additional step not present in Condition $N$, Agent C (Back-Translator) takes Agent B's English answer and translates it back into the original native language.

\end{itemize}

\textbf{Scoring Methodology:} To create an appropriate evaluation for all three conditions, we simulate the scenario faced by a deployed system when presenting its final output to the user in their native language by evaluating against the native-language gold answer. Condition E includes Agent C, as we want our metric to measure the entire end-to-end translation tax (forward translation, English reasoning, and back-translation). We intentionally include Agent C (back-translation) in Condition E because the 'English-Forcing Tax' is defined as the end-to-end, user-facing cost of deploying an English-pivot architecture. Omitting back-translation would evaluate an incomplete, non-deployable pipeline that cannot serve a non-English end-user. This avoids the scoring confound of evaluating English outputs against native-language gold answers, allowing the true architectural cost of routing through English to be measured accurately.

\subsection{Metrics and Statistical Procedure}
We measure performance using the standard SQuAD Exact Match (EM) and token-level F1 metrics. We use two complementary measures to separate the effects of multi-agent orchestration from cross-lingual translation.

First, we define the \textbf{Total Orchestration Penalty} relative to the single-agent baseline ($S$) as
\begin{equation}
\text{Total Penalty}
=
\text{Accuracy}(S)-\text{Accuracy}(c).
\end{equation}
Here, $c \in \{N, E\}$ denotes either the Native ($N$) or English-forced ($E$) multi-agent condition. This metric quantifies the overall performance degradation introduced by multi-agent execution, irrespective of the communication language.

Our primary metric for isolating the effect of English-forced communication is the \textbf{English-Forcing Tax}, defined as
\begin{equation}
\text{English-Forcing Tax}
=
\text{Acc.}(N)-\text{Acc.}(E).
\end{equation}
This metric quantifies the end-to-end cost of the English-pivot pathway encompassing forward-translation, English-language reasoning, and back-translation steps. Conditions $N$ and $E$ share the same extraction-answer core architecture (Agents A and B), while Condition $E$ adds an English-pivot translation pathway (forward translation, English processing, and Agent C back-translation) that Condition $N$ does not require. Conversely, the \textbf{Total Orchestration Penalty}, calculated as $\text{Accuracy}(S) - \text{Accuracy}(N)$ or $\text{Accuracy}(S) - \text{Accuracy}(E)$, evaluates the overall performance degradation relative to the single-agent baseline. Thus, $\text{Accuracy}(S) - \text{Accuracy}(E)$ measures both the overall cost of multi-agent orchestration and the additional degradation introduced by English-mediated communication, whereas $\text{Accuracy}(N) - \text{Accuracy}(E)$ isolates only the latter.

To rigorously test whether the English-Forcing Tax is statistically significant, we apply the exact two-sided McNemar's test \cite{dietterich1998approximate} to the paired binary EM outcomes (0 or 1) of Condition N versus Condition E for each question. We report the counts of discordant pairs ($b$ = Native correct/English wrong, $c$ = English correct/Native wrong). Rather than relying on the chi-square approximation, we compute the exact p-value using the binomial distribution $\mathcal{B}(n, 0.5)$ where $n = b + c$. The two-sided exact p-value is calculated as:
\begin{equation}
p = 2 \times \sum_{i=0}^{\min(b,c)} \binom{b+c}{i} \left(\frac{1}{2}\right)^{b+c}
\end{equation}
Since we conduct four multiple comparisons (one per language), we apply a conservative Bonferroni correction with a significance threshold of:
\begin{equation}
\alpha_{\text{corr}} = \frac{\alpha}{m} = \frac{0.05}{4} = 0.0125
\end{equation}

Moreover, to understand \textit{where} the pipeline fails in Condition E, we compute the character n-gram F-score (chrF) \cite{popovic-2015-chrf} between Agent A's forward translation and the independent English gold answer. This allows us to relate English-reference lexical overlap to final pipeline success while avoiding the circularity of scoring Agent C against the native-language gold answer.

\subsection{Compute and Reproducibility}
Every experiment was executed locally on a consumer-grade MacBook Air (M-series, 16GB RAM) using Apple's MLX framework to demonstrate the accessibility of this research and the practicality of deploying agent systems on edge devices. The entire experiment (3,600 pipeline executions corresponding to approximately 7,200 LLM generations) was completed sequentially in about 8 hours. The exact prompt templates for Agents A, B, and C are provided in Appendix~\ref{sec:appendix_prompts}.

To ensure the deterministic outputs required for rigorous paired statistical testing (McNemar's test), all LLM generations were performed using greedy decoding (temperature = 0.0). We acknowledge that evaluating temperature variance and sampling stochasticity in multi-agent pipelines remains an important avenue for future work

\section{Results}

The main hypothesis has solid statistical support. Our primary comparison is the English-Forcing Tax ($N - E$), which quantifies the end-to-end cost of the English-pivot pathway (including forward-translation and back-translation steps) while keeping the underlying multi-agent architecture fixed. Across all four languages, the English-forced condition consistently exhibits a larger performance degradation than the native multi-agent condition. We also report the Total Orchestration Penalty relative to the single-agent baseline to measure the overall end-to-end cost of multi-agent orchestration. Together, these metrics separate the general cost of orchestration from the additional degradation introduced by the English translation pathway.

\begin{table*}[t]
\centering
\small
\setlength{\tabcolsep}{4pt}
\begin{tabular}{clcccccc}
\toprule
\textbf{Lang} & \textbf{Cond.} & \textbf{EM} & \textbf{F1} & \textbf{Pen. (EM)} & \textbf{Pen. (F1)} & \textbf{$b$, $c$} & \textbf{$p$-value} \\
\midrule
\multirow{3}{*}{\textbf{es}}
& S & 0.637 & 0.840 & -- & -- & -- & -- \\
& N & 0.553 & 0.733 & 0.083 & 0.107 & \multirow{2}{*}{66, 27} & \multirow{2}{*}{$6.47 \times 10^{-5}$} \\
& E & 0.423 & 0.620 & \textbf{0.130} & \textbf{0.113} & & \\
\midrule
\multirow{3}{*}{\textbf{ar}}
& S & 0.547 & 0.752 & -- & -- & -- & -- \\
& N & 0.430 & 0.632 & 0.117 & 0.120 & \multirow{2}{*}{67, 17} & \multirow{2}{*}{$3.50 \times 10^{-8}$} \\
& E & 0.263 & 0.464 & \textbf{0.167} & \textbf{0.168} & & \\
\midrule
\multirow{3}{*}{\textbf{zh}}
& S & 0.627 & 0.828 & -- & -- & -- & -- \\
& N & 0.510 & 0.721 & 0.117 & 0.106 & \multirow{2}{*}{102, 19} & \multirow{2}{*}{$6.37 \times 10^{-15}$} \\
& E & 0.233 & 0.540 & \textbf{0.277} & \textbf{0.181} & & \\
\midrule
\multirow{3}{*}{\textbf{hi}}
& S & 0.570 & 0.740 & -- & -- & -- & -- \\
& N & 0.443 & 0.596 & 0.127 & 0.145 & \multirow{2}{*}{106, 14} & \multirow{2}{*}{$1.16 \times 10^{-18}$} \\
& E & 0.137 & 0.317 & \textbf{0.306} & \textbf{0.279} & & \\
\bottomrule
\end{tabular}
\caption{Main results for Exact Match (EM) and F1 scores evaluated on 300 samples per language. \textbf{Pen.} decomposes the performance loss: for Condition N, it reports the baseline Multi-Agent Overhead ($S - N$); for Condition E, it reports the primary result - the \textbf{English-Forcing Tax} ($N - E$) which isolates the added cost of routing inter-agent communication through English. Here, $b$ and $c$ represent the discordant pairs ($b$ = Native correct/English wrong, $c$ = English correct/Native wrong). All reported $p$-values are derived from McNemar's test comparing Conditions N and E and remain significant after Bonferroni correction ($\alpha = 0.0125$).}
\label{tab:main_results}
\end{table*}

\begin{figure}[t]
    \centering
    \includegraphics[width=\linewidth]{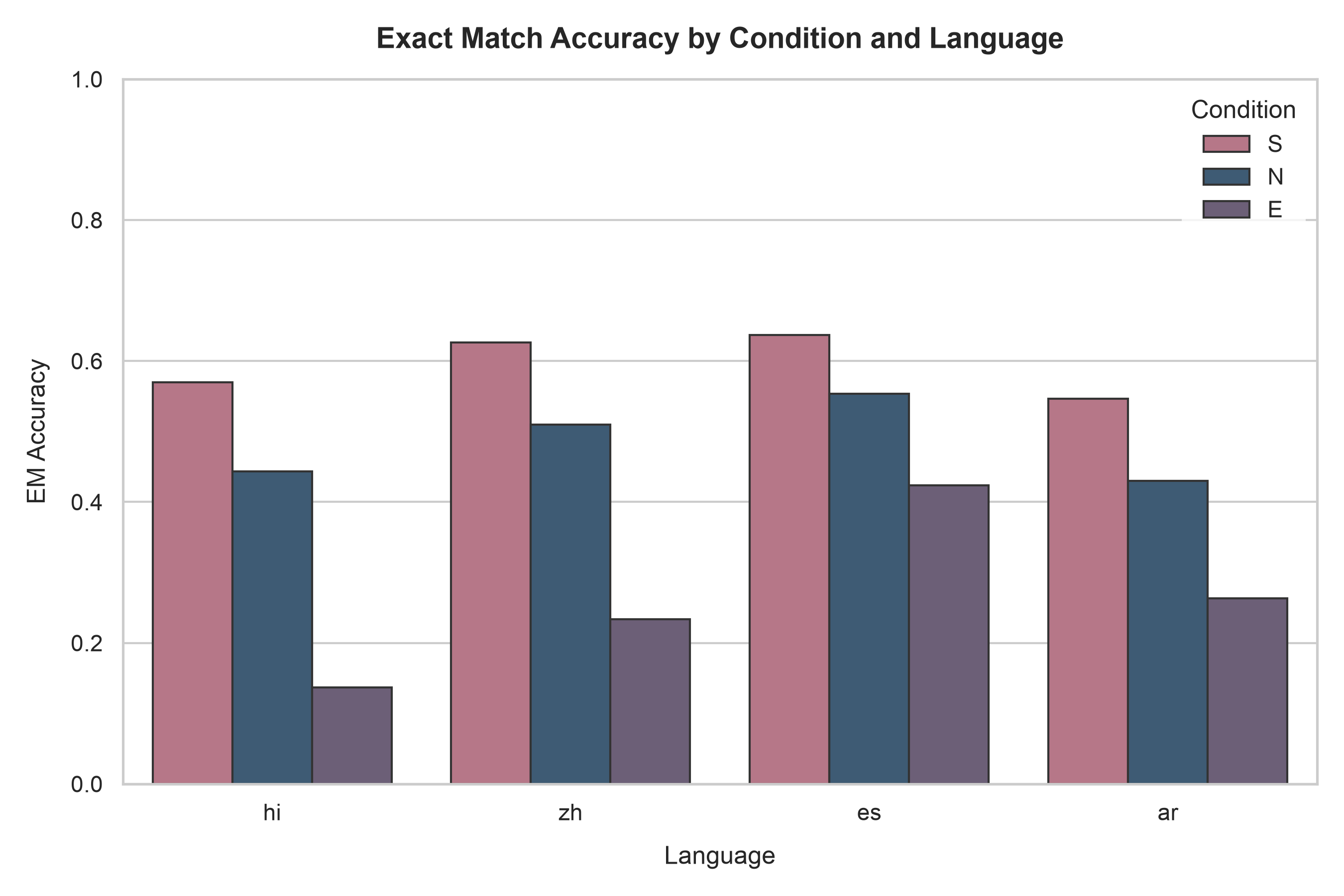}
    \caption{Exact Match (EM) accuracy for each language across the Single-Agent (S), Native Multi-Agent (N), and English-Forced Multi-Agent (E) conditions. All differences between Conditions N and E are statistically significant after Bonferroni correction ($\alpha = 0.0125$).}
    \label{fig:em_accuracy}
\end{figure}

\begin{figure}[t]
    \centering
    \includegraphics[width=\linewidth]{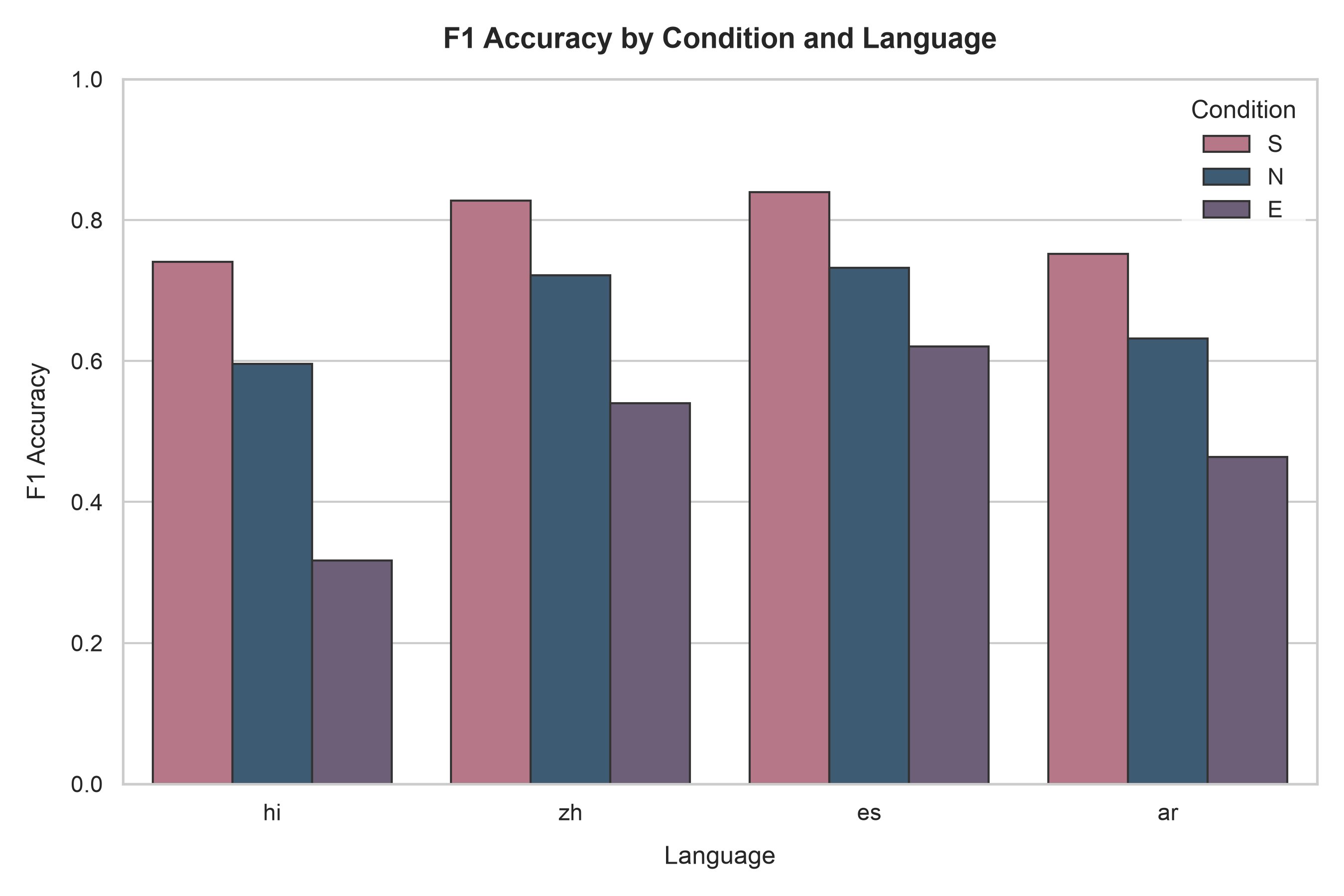}
    \caption{F1 scores under the same conditions. In contrast to Hindi, where the English-Forcing Tax ($N-E$) for EM and F1 closely agree (0.306 vs. 0.279), the F1 penalty for Chinese is meaningfully smaller than its EM penalty (0.181 vs. 0.277), indicating greater partial token overlap in Chinese failure cases compared to Hindi.}
    \label{fig:f1_accuracy}
\end{figure}

\textbf{Typological Gradient and Statistical Significance.}
 Table~\ref{tab:main_results} shows that the incremental English-Forcing Tax ($N - E$) increases substantially with linguistic distance from English. Spanish, a Romance language that uses the Latin script like English, incurs the smallest English-forcing penalty (13.0 percentage-point EM drop; 11.3 percentage-point F1 drop). By comparison, more typologically distant languages experience much greater degradation: Arabic declines by 16.7 percentage points, while Chinese (27.7 percentage points) and Hindi (30.6 percentage points) incur the largest accuracy penalties from routing through English. Although this pattern ($es < ar < zh < hi$) is roughly consistent with other typological distance measures as well as distance from the English script, we note that it is also confounded with per-language data size and resource availability in the base model. Future cross-lingual agent research should disentangle strict typological distance from training-data resourcedness. Crucially, all four pairwise comparisons ($N$ vs.\ $E$) remain statistically significant after a conservative Bonferroni correction ($\alpha = 0.0125$), with p-values ranging from $6.47 \times 10^{-5}$ (Spanish) to $1.16 \times 10^{-18}$ (Hindi).

\textbf{Discordant Pair Asymmetry.}
The discordant pairs ($b$, $c$) from McNemar's test reveal the systematic nature of this information loss. For Hindi, there are 106 instances where the Native pipeline succeeds but the English-forced pipeline fails, compared to only 14 instances of the reverse. This asymmetry follows the same typological trend across languages: Spanish (66:27), Arabic (67:17), Chinese (102:19), and Hindi (106:14). These highly asymmetric discordant pairs confirm that the English-forced pipeline consistently converts correct predictions into incorrect ones rather than merely introducing random variation.

\begin{figure}[t]
    \centering
    \includegraphics[width=\linewidth]{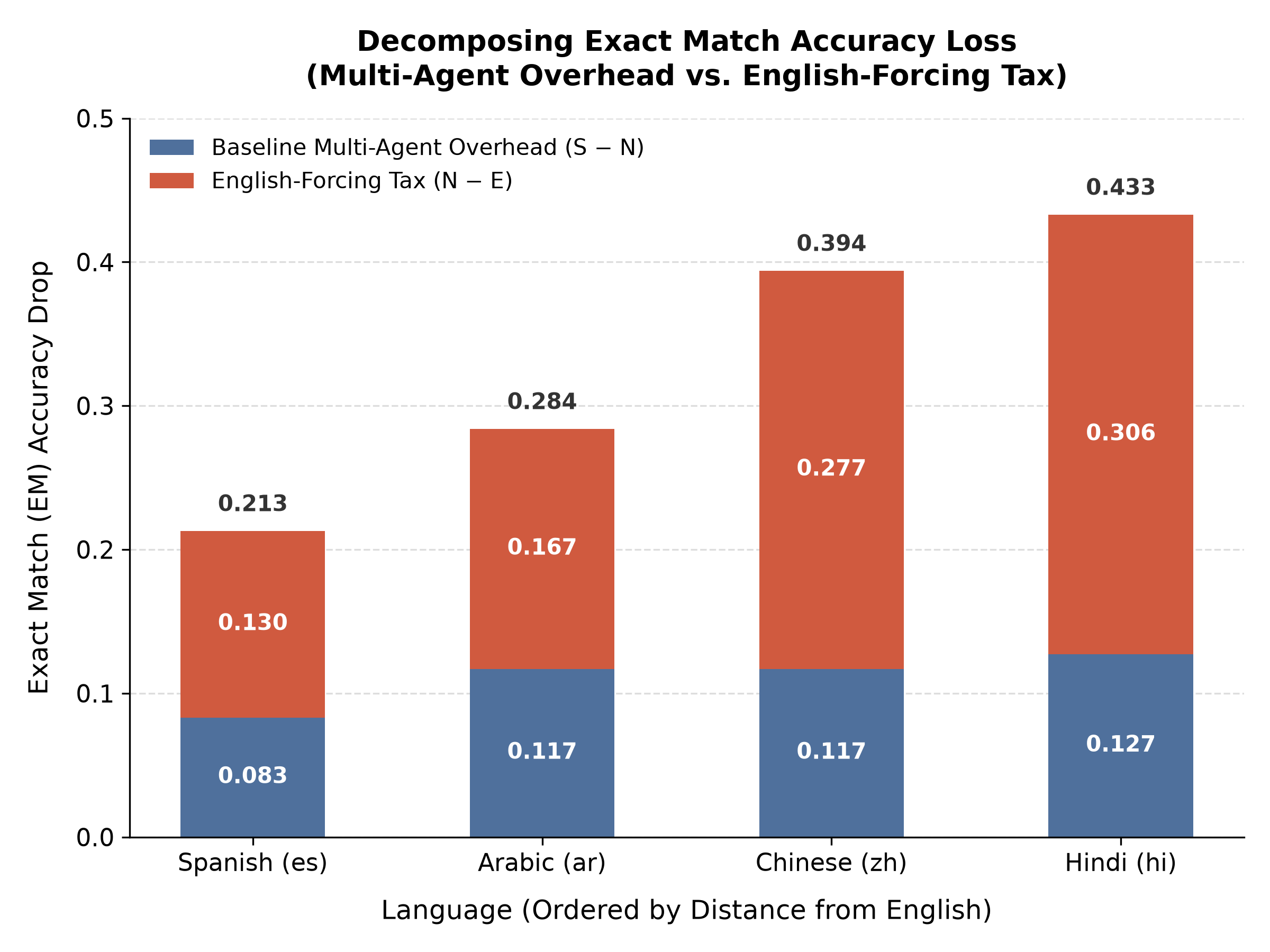}
    \caption{Decomposition of Exact Match (EM) accuracy loss across languages. The total height of each bar represents the overall drop from the Single-Agent baseline to the English-Forced condition ($S - E$). This loss is decomposed into the baseline \textbf{Multi-Agent Overhead} ($S - N$, bottom blue segment) and our primary finding, the \textbf{English-Forcing Tax} ($N - E$, top orange segment), which isolates the incremental accuracy drop caused by routing inter-agent communication through English.}
    \label{fig:penalties}
\end{figure}

\begin{figure}[t]
    \centering
    \includegraphics[width=\linewidth]{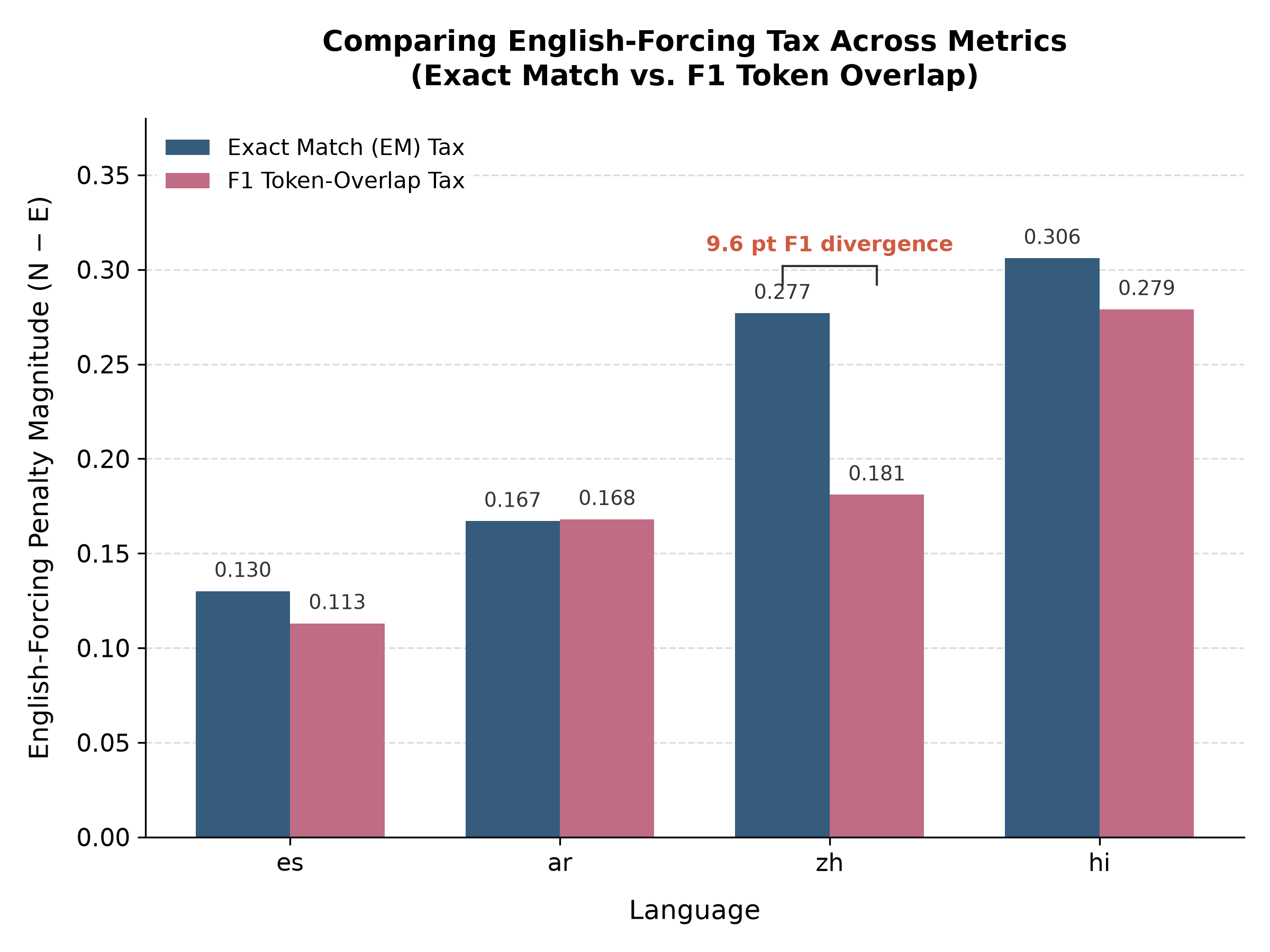}
    \caption{Comparison of the \textbf{English-Forcing Tax} ($N - E$) measured by Exact Match (EM) versus F1 token-overlap accuracy. For Hindi, Arabic, and Spanish, the EM and F1 penalties align closely, indicating that incorrect outputs rarely retain partial n-gram correspondence with the reference answer. In contrast, Chinese exhibits a \textbf{9.6 percentage-point F1 divergence} ($0.277$ EM drop vs.\ $0.181$ F1 drop), indicating greater partial meaning or token-level overlap in failure cases.}
    \label{fig:em_vs_f1_penalty}
\end{figure}

\textbf{The Chinese F1 Divergence.}
Analysis of the F1 scores reveals a distinct, language-dependent failure mode. For Hindi, both the English-Forcing Tax EM (0.306) and F1 (0.279) penalties correlate closely, indicating that incorrect outputs rarely have even partial word-level correspondence with the reference answer. In contrast, for Chinese, the F1 penalty (0.181) is significantly smaller than the EM penalty (0.277).
 The 9.6 percentage-point difference indicates that corrupted Chinese outputs retain some meaning-level overlap with the gold reference and do not suffer simply from a loss of contextual keywords, but may instead result from lexical variation or paraphrasing. Although this difference should be investigated further through human evaluation or semantic similarity scores, it does show that the English-Forcing Tax appears to operate differently at the token level across languages.

\section{Diagnostic Correlational Analysis of English-Reference Lexical Overlap}

The results in Section 4 suggest a substantial accuracy cost incurred by routing agent handoffs through English. However, Condition E combines three possible sources of error: Agent A's forward translation, Agent B's English reasoning, and Agent C's back-translation. To derive actionable engineering implications, we need to isolate \textit{which} step is responsible for the failure.

For example, if the loss is primarily caused by Agent B reasoning poorly in English, then the solution would be to improve the underlying model's English task capabilities. On the other hand, if the loss originates from the translation steps, then a more appropriate architectural solution is to use native-language routing, bypassing the translation bottleneck entirely. To do this, we performed a diagnostic correlational analysis of the lexical overlap between Agent A's English output and the English reference to examine the final pipeline outcome.

\begin{table}[t]
\centering
\small
\resizebox{\columnwidth}{!}{%
\begin{tabular}{lcc}
\toprule
\textbf{Pipeline Outcome} & \textbf{Mean Agent A chrF} & \textbf{Sample Count ($n$)} \\
\midrule
Success (EM = 1) & 83.90 & 322 \\
Failure (EM = 0) & 56.97 & 878 \\
\bottomrule
\end{tabular}%
}
\caption{Correlational diagnostics for Condition E failure. Agent A's English-reference lexical overlap (chrF) was evaluated against a separate English gold reference. Lower lexical overlap is more strongly associated with downstream pipeline failure.}
\label{tab:chrf_decomposition}
\end{table}

\begin{figure}[t]
    \centering
    \includegraphics[width=\linewidth]{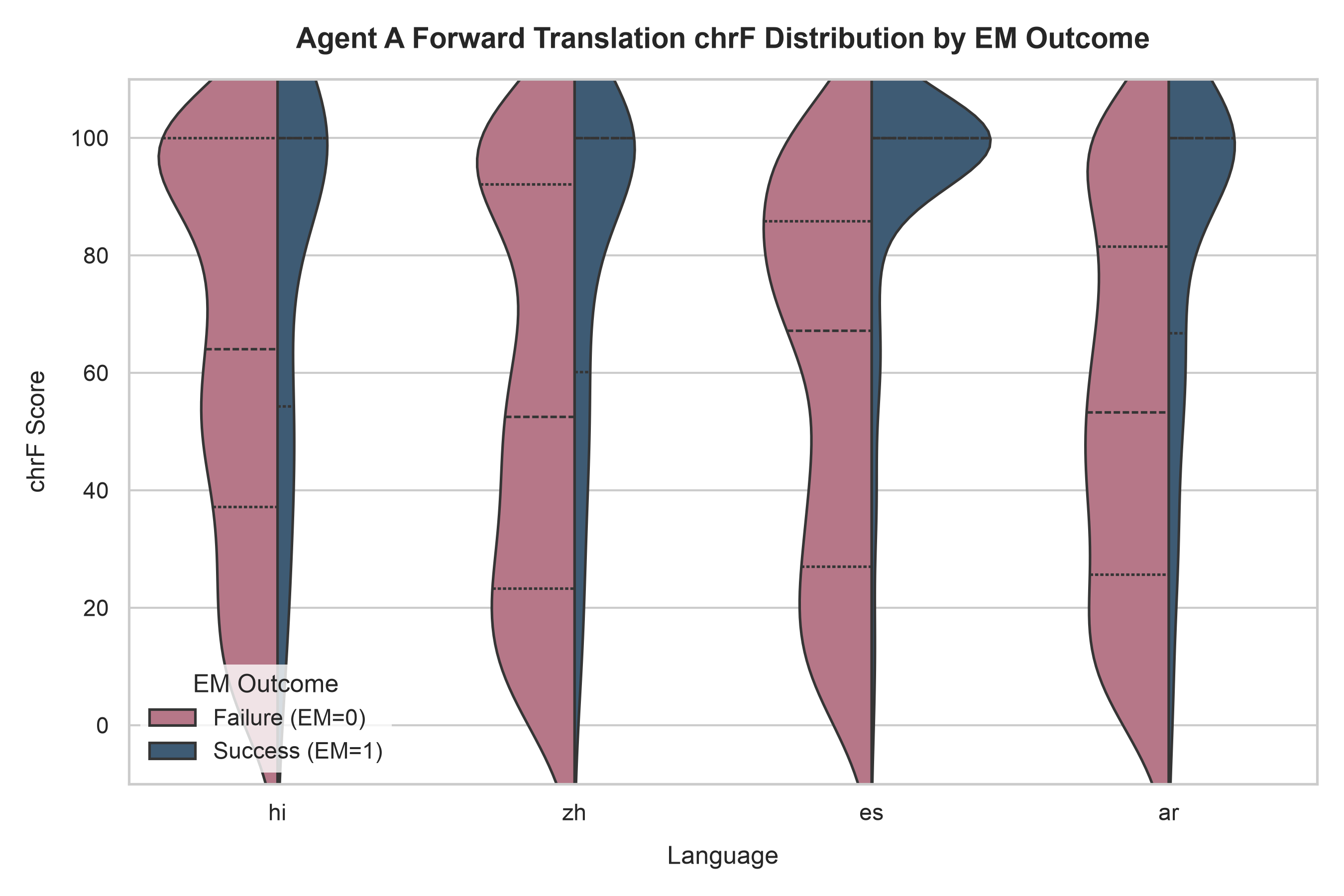}
    \caption{Agent A's English-reference lexical overlap (chrF), scored against an independent English gold reference, by final pipeline outcome. Predictions with lower lexical overlap are strongly associated with pipeline failure (mean chrF 56.97 for failures vs. 83.90 for successes, $n=878/322$).}
    \label{fig:chrf_decomposition}
\end{figure}

We calculate the character n-gram F-score (chrF) \cite{popovic-2015-chrf} between Agent A's English output and the English gold answer for each of the 1,200 Condition E executions. We stress that this is a correlational diagnostic measure of lexical overlap rather than a strict translation fidelity score, as it compares the extracted facts to the final gold answer. We then correlate this lexical overlap score with the final pipeline result (Success or Failure).

As shown in Table~\ref{tab:chrf_decomposition} and Figure~\ref{fig:chrf_decomposition}, the correlation is clear.  When the pipeline succeeds, Agent A's English output has substantially higher lexical overlap with the English gold answer (mean chrF = 83.90) than when the pipeline fails (mean chrF = 56.97).

Lower English-reference lexical overlap was identified as a strong correlate of pipeline failure. Although this diagnostic metric does not fully isolate translation errors from downstream English reasoning errors, the strong correlation suggests that when Agent A cannot faithfully translate the extracted facts, this failure will often precede and likely contribute to failures in the subsequent reasoning stage. Since Agent B only receives the noisy English text, it often loses access to the correct semantic entities before compounding hallucinations, as shown by the entity mistranslation examples in Section 6. The engineering takeaway is therefore compelling: while optimizing English reasoning remains important, agent frameworks should strongly consider supporting native-language routing so that the lossy forward-translation step can be skipped entirely.

\section{Illustrative Failure Modes}

To provide qualitative insight into the mechanisms underlying the quantitative loss, we manually examined 20 cases where Condition N succeeded but Condition E failed ($b$ discordant pairs). Although not intended as a statistically rigorous taxonomy, four recurring illustrative failure modes emerged:

\begin{itemize}

\item \textbf{Entity and Domain-Specific Mistranslation:} Agent A mistranslates a proper noun or technical term, preventing Agent B from arriving at the correct answer that Condition N successfully produced. For example, the NFL player ``Kony Ealy'' was mistranslated as ``Be Like Eli'', leading Agent B to confidently hallucinate ``Eli Manning''.

\item \textbf{Semantic Drift and Compounding Hallucination:} While Condition N preserves the native context, the shift to English in Condition E causes Agent B to lose nuance or over-compress the information. For example, Agent A correctly extracted ``Revolutionary Civil Disobedience'', but Agent B over-compressed it to simply ``Revolution''.

\item \textbf{Local Entity Dropping:} Agent A drops local terms without a direct English equivalent (e.g., certain administrative terminology) to simplify the translation, sacrificing important contextual information during the handoff that Condition N retained.

\item \textbf{Paraphrase-Induced Penalties:} The pipeline preserves the underlying concept, but multi-step translation introduces lexical noise (e.g., ``Sierra Freeway'' becomes ``Sierra Highway''), breaking strict Exact Match scoring without a complete loss of semantic meaning.

\end{itemize}

\section{Discussion}

The dominant engineering belief that routing multilingual agent handoffs through English leads to better reasoning carries a hidden, compounding translation cost. For typologically distant languages, our results provide strong evidence that this architectural choice sacrifices up to 30.6 percentage points of Exact Match accuracy relative to a native-language multi-agent baseline. Moreover, since degraded forward translation is strongly associated with pipeline failure, the engineering implication is clear: optimizing English reasoning alone is unlikely to be sufficient. Instead, agent frameworks should support native-language routing throughout the orchestration layer. Our results build on recent empirical evidence that even architectural overhead can become a bottleneck for multi-agent systems. Although previous work evaluated the diminishing value of adding more agentic computation \citep{agrawal2026token}, here we demonstrate that indirect language routing through repetitive cross-lingual translation incurs a separate cost to orchestration.

\section{Conclusion}

The standard engineering assumption that routing multilingual agent handoffs through English leads to better reasoning imposes a significant hidden translation cost. This work isolates the architectural contribution of this ``English-internals'' default by evaluating a controlled two-agent pipeline across four typologically diverse languages. We demonstrate that forcing inter-agent communication through a single end-to-end English translation bottleneck incurs a statistically significant English-Forcing Tax, sacrificing up to 30.6 percentage points of Exact Match accuracy compared with native-language orchestration for typologically distant languages such as Hindi and Chinese. Additionally, our diagnostic analysis reveals a strong relationship between pipeline failure and reduced English-reference lexical overlap due to lossy forward translation, consistent with our finding that translation loss is an important contributor to the observed performance drop. These results provide strong evidence that English-forced agent pipelines introduce additional barriers to global language access. To build robust and equitable multilingual AI systems, developers should redesign agent frameworks to move beyond the English-internals default and support native-language inter-agent communication.

\section*{Limitations}

Using a single, high-capacity (8B) multilingual model and a controlled extraction-answer pipeline with a two-agent core and an additional back-translation agent in the English-forced condition, we reduce confounding factors. An 8-bit quantized version of Aya-23-8B is also used for consumer-grade edge deployment. Prior work has shown that quantization can introduce performance degradation in LLMs and that sensitive weights may be disproportionately affected by quantization-induced perturbations \cite{li2026semq}. Quantization may therefore influence the magnitude of the observed tax, but the direction of this effect cannot be determined without an unquantized control. We leave the analysis of larger, unquantized frontier models, which may exhibit different failure modes, to future API-based investigations. Our pipeline also relies entirely on zero-shot prompting. Lexical overlap may be improved with few-shot examples or more advanced prompt engineering, but zero-shot remains the out-of-the-box default for many widely used agentic frameworks such as AutoGen and LangChain. We concentrate on measuring the baseline architectural tax from established default paradigms, while leaving prompt-engineered mitigation strategies for future work. Twenty failure cases are presented as a qualitative illustration rather than a validated taxonomy. Evaluating additional model families (e.g., Llama-3, Mistral), parameter scales (7B vs. 70B), and deeper agent chains with more than three hops should be part of future research.

\section*{Acknowledgments}
This work was supported by the Natural Science Foundation of Chongqing (Grant No. CSTB2024NSCQ-LZX0083).


\bibliography{custom}

\appendix
\onecolumn
\section{Appendix A: Prompt Templates}
\label{sec:appendix_prompts}

To ensure full reproducibility, we provide the exact prompt templates used for each agent in our pipeline. All prompts utilize the model's chat template (via the \texttt{apply\_chat\_template} function in Hugging Face/MLX). \texttt{\{lang\_name\}} is dynamically replaced with the target language (e.g., Hindi, Chinese, Spanish, Arabic). The variables \texttt{\{context\}}, \texttt{\{question\}}, and \texttt{\{english\_question\}} are populated from the aligned XQuAD dataset, where \texttt{\{english\_question\}} denotes the corresponding English question for the same example. In the multi-agent setting, Agent A performs a single native-language fact extraction that is shared by both Conditions N and E.

\textbf{Condition S (Single-Agent Baseline):}
\begin{prompt}
Read the following passage in {lang_name}. Answer the question in {lang_name}.

Passage: {context}
Question: {question}

Answer:
\end{prompt}

\textbf{Condition N (Native Multi-Agent):}
\begin{prompt}
# Agent A (Shared Extractor)
Read the following passage in {lang_name}. Extract the facts relevant to
answering the question. Output ONLY the extracted facts in {lang_name}.

Passage: {context}
Question: {question}

Facts:

# Agent B (Answerer)
Based ONLY on the following facts, answer the question in {lang_name}.

Facts: {agent_a_output}
Question: {question}

Answer:
\end{prompt}

\textbf{Condition E (English-Forced Multi-Agent):}
\begin{prompt}
# Shared Agent A Output
(The native-language facts produced by Agent A above are reused.)

# Agent A - Translate to English
Translate the following text into English. Output ONLY the English
translation, with no extra text.

Text: {agent_a_output}

English Translation:

# Agent B (Answerer)
Based ONLY on the following English facts, answer the question in English.

Facts: {agent_a_english_facts}
Question: {english_question}

Answer:

# Agent C (Back-Translator)
Translate the following English text into {lang_name}. Output ONLY the
{lang_name} translation, with no extra text.

Text: {agent_b_english_output}

{lang_name} Translation:
\end{prompt}

\end{document}